\documentclass{article}
\usepackage{spconf,amsmath,graphicx}

\usepackage{hyperref}
\usepackage{soul}
\usepackage{amsmath}
\usepackage{amsthm}
\usepackage{booktabs}
\usepackage{amsfonts,amssymb}
\usepackage{multirow}
\usepackage{subfigure}
\usepackage{makecell}
\newcommand{\tabincell}[2]{\begin{tabular}{@{}#1@{}}#2\end{tabular}}

\title{Text Classification in the Wild: \\A Large-Scale Long-Tailed Name Normalization Dataset}
\name{Jiexing Qi,  Shuhao Li,  Zhixin Guo,  Yusheng Huang, 
Chenghu Zhou, Weinan Zhang, Xinbing Wang, Zhouhan Lin
\thanks{Zhouhan Lin is the corresponding author.}
}
\address{Shanghai Jiao Tong University, Shanghai, China}
\begin{document}
%
\maketitle
\begin{abstract}
Real-world data usually exhibits a long-tailed distribution, with a few frequent labels and a lot of few-shot labels. The study of institution name normalization is a perfect application case showing this phenomenon. There are many institutions worldwide with enormous variations of their names in the publicly available literature. 
In this work, we first collect a large-scale institution name normalization dataset \textbf{LoT-insts\footnote{\textbf{Lo}ng-\textbf{T}ailed \textbf{inst}ituition name\textbf{s}}}, which contains over 25k classes that exhibit a naturally long-tailed distribution. In order to isolate the few-shot and zero-shot learning scenarios from the massive many-shot classes, we construct our test set from four different subsets: many-, medium-, and few-shot sets, as well as a zero-shot open set. 
We also replicate several important baseline methods on our data, covering a wide range from search-based methods to neural network methods that use the pretrained BERT model.
Further, we propose our specially pretrained, BERT-based model that shows better out-of-distribution generalization on few-shot and zero-shot test sets. 
Compared to other datasets focusing on the long-tailed phenomenon, our dataset has one order of magnitude more training data than the largest existing long-tailed datasets and is naturally long-tailed rather than manually synthesized. We believe it provides an important and different scenario to study this problem. 
To our best knowledge, this is the first natural language dataset that focuses on long-tailed and open-set classification problems.\footnote{We release our code at \url{https://github.com/LUMIA-Group/LoT-insts}. The dataset website can be found at \url{https://lot-insts.github.io/}}
\end{abstract}


\section{Introduction}
Real-world data often have a long-tailed distribution \cite{liu2019large}. For example, the frequency of words in natural language \cite{newman2005power}, the number of connections social media users have \cite{weller2016trying}, the species richness in an ecosystem \cite{albert2011historical}, etc. Also, real-world data distributions are open-ended: new classes may show up in the actual world. Therefore, applying classification or recognition models to a real-world case is nontrivial from a scientific point of view: scenarios of few-shot and zero-shot learning will be encountered.

\begin{table}[t]
\small
\begin{center}
\begin{tabular}{c}
\toprule
\tabincell{c}{Name Instances for \texttt{Yale University}} \\
\midrule
\tabincell{l}{Yale Univ, New Haven, CT}      \\  
\tabincell{l}{Yale\#N\# \quad \quad     University}  \\ 
\tabincell{l}{Yale Medical School, New Haven, Connecticut4}      \\ 
\tabincell{l}{Yale U., Phys Dept., 217 Prospect St., New Haven, CT 06511.}      \\ 
\bottomrule
\end{tabular}
\end{center}
\caption{An example illustrating some different variations extracted from academic publications that correspond to a same normalized name (Yale University, in this example).}
\label{tab:sample_demo}
\end{table}

\begin{table}[t]
    \centering
    \begin{tabular}{c|c|c|c|c}
\toprule
Dataset & \makecell{ImgNet\\-LT} & \makecell{Places\\-LT} & \makecell{MS1M\\-LT} & \makecell{LoT-insts}   \\ 
\midrule
Classes        & 1,000         & 365        & 74,532      & 25,129           \\ 
Train          & 115,846       & 62,500      & 887,530     & 2,238,148        \\ 
Valid          & 20,000        & 7,300       & -          & 54,439               \\ 
Test           & 50,000        & 36,500      & 3,530      & 58,154          \\ 
Max            & 1,280         & 4,980       & 598        & 22,606          \\ 
Min            & 5            & 5          & 1          & 1               \\ 
Dist.          & Rsmpl.    & Rsmpl.  & Rsmpl.  & Natural         \\ 
\bottomrule
\end{tabular}
\caption{Comparison between our dataset and others. The other three datasets are from \cite{liu2019large}. \textit{Rsmpl.} is short for \textit{Resampled}, \textit{LT} is short for \textit{Long-tailed}, and \textit{ImgNet} for \textit{ImageNet}. }
\label{dataset_comparison}
\end{table}

In recent years, there has been increasing interest in the study of long-tailed data, but mostly within the field of computer vision. Since \cite{liu2019large} presented a visual classification dataset by resampling labeled images from ImageNet \cite{deng2009imagenet}, related research on this specific scenario has been greatly motivated. Such as 
through resampling training sets (Decouple \cite{kang2019decoupling}, BBN \cite{zhou2020bbn}), reweighting loss function \cite{cui2019class, cao2019learning, jamal2020rethinking}, or with transfer learning \cite{liu2020deep, xiang2020learning}. To our best knowledge, there are no public datasets for long-tailed classification in the field of NLP, thus hindering the development of related techniques for natural language. 

Name normalization for academic institutions is a text classification task that classifies non-standard institution names into classes consisting of their standard forms. The non-standard institution names are usually extracted by OCR or PDF parsing algorithms on academic publications, which could be written in different granularity (department/school/university), abbreviations (MIT/M.I.T.), use of accent characters or typographical errors, etc. (See Table \ref{tab:sample_demo}). These lexical variations of institution names result in redundancy, incompleteness, and ambiguity \cite{fatma2020canonicalizing}, which poses serious problems for a bunch of downstream tasks, such as information retrieval or author profiling.

In this work, we construct a large dataset of academic institutions named \textbf{LoT-insts} for the name normalization task from Microsoft Academic Graph (MAG) \cite{sinha2015overview} and present it as a text classification task, with the extracted affiliation names being examples and their corresponding deduplicated institution names being classes.
We present \textbf{LoT-insts} in a form that emphasizes its long-tailed distribution nature of it by isolating zero-shot, few-shot sets out from medium- and many-shot sets in the test data. Different from other similar computer vision datasets such as those sampled from ImageNet, Places, or MS1M-ArcFace datasets \cite{liu2019large}, our dataset has far more classes and several orders of magnitudes more instances (see Table \ref{dataset_comparison}). In addition to that, our data are naturally long-tail distributed rather than manually sampled after collection, thus reflecting a more authentic distributional nature of real-world data. To our best knowledge, this is the first natural language dataset that focuses on this problem. 

Considering that de-duplication of institutions in new classes is of practical importance, we propose a new open-set evaluation task named \textit{open-set class verification (OSV)}, which verifies if two given examples are of the same never-seen class or not.

We further reproduce several baseline methods of important previous works on name normalization, including Naive Bayes \cite{maron1961automatic}, sCool \cite{jacob2014scool}, and CompanyDepot \cite{liu2016companydepot}. These methods not only provide a first comparative evaluation of them on the same dataset but also provide an analysis of their performances in the frequent and non-frequent classes. 
In addition to reproducing existing models, we also propose a BERT-based, pretrained model on this large dataset, which shows better performance across many- and few-shot test sets.
We believe these contributions combined could pave the first stone of the way towards long-tailed text classification.

\section{Related Work}
\begin{figure*}[!t]
    \centering
    \includegraphics[width=2.0\columnwidth]{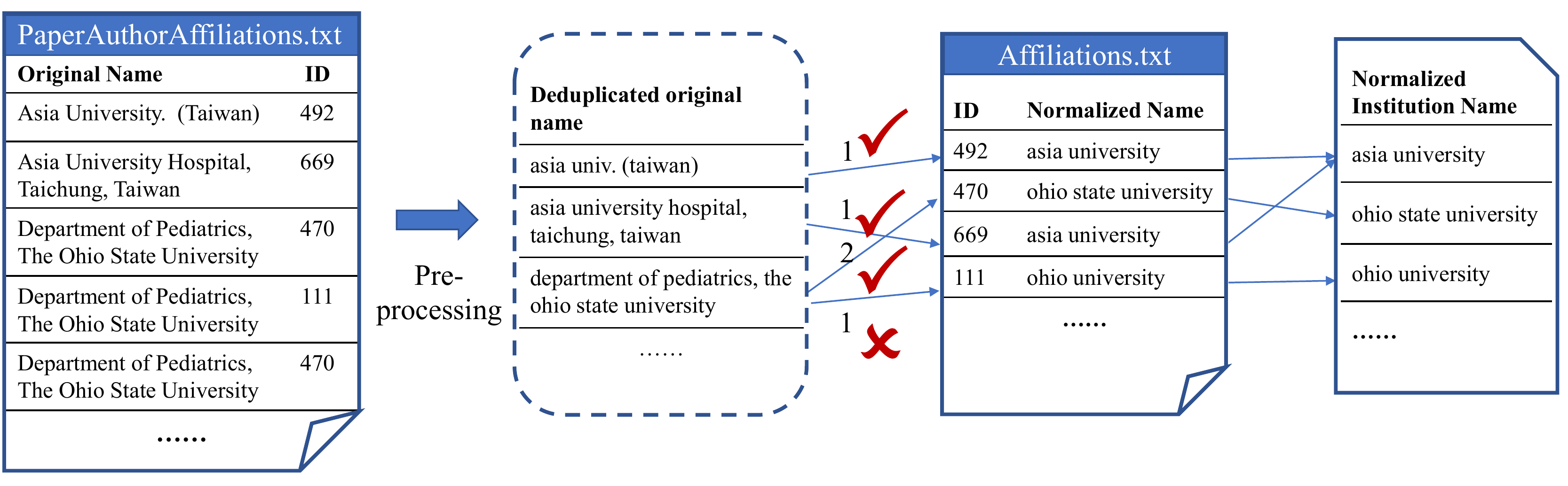}
    \caption{Process of data cleaning during data collection. We first extract original institution names and institution IDs from the PaperAffiliations.txt file (leftmost). With preprocessing, we clean the data by removing noise from the original files and then deduplicate original names. Thus we get a record noting the occurrence of each mapping (middle-left). We then detect and remove wrong mappings and remove ambiguous mappings by using some voting method (middle-right). Finally, we clean the institution IDs by removing duplicated institutions. Thus we acquire high-quality mappings from clean non-normalized names to the clean IDs of their normalized names.}
    
    \label{fig:extractdata}
\end{figure*}

There are generally three categories that aim at tackling the open long-tailed classification since it is first formulated in  \cite{liu2019large} as an image recognition task. The first is resampling, which is the reverse weighting of the sampling frequency of different categories of images according to the number of samples. For example,  \cite{shen2016relay,buda2018systematic,byrd2019effect} are methods around oversampling the few-shot classes, while  \cite{buda2018systematic,japkowicz2002class} propose methods around undersampling the abundant data from dominant classes. Repeating tail samples may lead to overfitting of a few classes \cite{chawla2002smote,cui2019class} while discarding valuable data will inevitably affect the generalization ability of the learned model  \cite{zhou2020bbn}. The second category is reweighting, which focuses on increasing the loss function contribution of the tail categories. Usually, a large weight is allocated to the training samples of the tail class in the loss function  \cite{huang2016learning}. However, reweighting cannot deal with real large-scale scenes with long-tail data, which often leads to optimization difficulties  \cite{mikolov2013distributed}. The third one is transfer learning, which learns the general knowledge from the head classes, and then transfers the learned knowledge to the tail classes  \cite{liu2020deep,xiang2020learning}. Note that all of the methods above are proposed under the image recognition background, while there is a lack of related research in NLP due to the absence of a natural language dataset reflecting this long-tailed distribution phenomenon. 

As for the task of name normalization, there are generally three kinds of approaches. The first is rule-based methods, which use a set of expert-defined rules to categorize text into different classes, such as NEMO  \cite{jonnalagadda2010nemo}, and name normalization for product items  \cite{borkovsky2003item}, genes  \cite{wermter2009high}, diseases  \cite{leaman2013dnorm}, and persons  \cite{magdy2007arabic}, etc. 
Secondly, methods that rely on the hand-crafted features with classical machine learning models, such as Naive Bayes  \cite{maron1961automatic}.
And lastly, methods that use an external knowledge base. This type of method retrieves candidates from a knowledge base and ranks the outcomes from heuristics. Typical systems of this type include sCool  \cite{jacob2014scool} and CompanyDepot  \cite{liu2016companydepot}. 

More broadly, text classification is a well-studied topic that our task belongs to, albeit mostly in small numbers of classes, such as sentiment classification  \cite{maas-EtAl:2011:ACL-HLT2011}, topic classification  \cite{journals/semweb/LehmannIJJKMHMK15}, and natural language inference  \cite{marelli2014semeval}. Almost all of these datasets are well-balanced and contain small numbers of classes, which are not suited to studying the few-shot and zero-shot learning problems. Hierarchical text classification (HTC), on the other hand, usually involves a large number of classes with a tree-structured taxonomy. Datasets in this category involves LSHTC \cite{partalas2015lshtc}, BIOASQ \cite{tsatsaronis2015overview} and WikiSeeAlsoTitles-350K \cite{kharbanda2021embedding} etc. Various methods have been proposed around these works, focusing on leveraging the structural information between classes, but researchers on this line are less concerned with the few-shot scenario and don't have a zero-shot setting for out-of-distribution (OOD) test samples. 
A great variety of deep neural network-based models have been proposed for these tasks, and most recently, the state-of-the-art comes with a pretrained model, such as BERT  \cite{DBLP:conf/naacl/DevlinCLT19}, or ELMO  \cite{peters2018deep}.

\section{Dataset Construction}

We collect our dataset from Microsoft Academic Graph (MAG) \cite{sinha2015overview}, which is a heterogeneous graph containing scientific publication records, citation relationships between those publications, as well as authors, institutions, journals, conferences, fields of study, etc. There are mainly three stages in collecting our dataset from MAG. The first stage is data cleaning, where we remove most of the labeling noises from MAG. The second is data filtering, which removes redundant examples. Finally, we partition the dataset for training and evaluation.

\subsection{Data Cleaning}
\label{dataset_clean}

\begin{figure*}
    \centering
    \includegraphics[width=2\columnwidth]{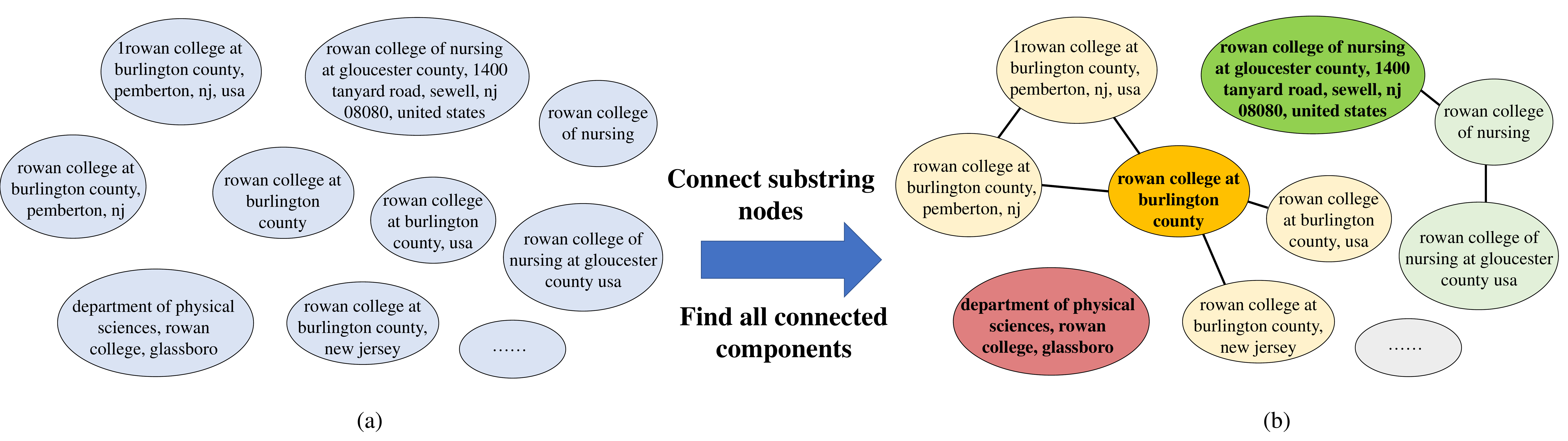}
    \caption{Example of data filtering to eliminate the redundant examples. (a) shows some original institution names that correspond to Rowan University. In (b), for each example, we detect if it is a substring of another within the examples of the same class. If so, we connect the two examples with an edge. This process forms an undirected graph with those examples being its nodes. For each connected component in the undirected graph, we randomly retain only one example (darker colored), and all others are discarded (shallower colored). }
    \label{fig:connected}
\end{figure*}

In the data cleaning stage, we extract a large set of mappings from non-standard, original institution names to their normalized ones. The overall procedures in this step are outlined in Figure \ref{fig:extractdata}. 

Two files are utilized in extracting our dataset from MAG. The first is \textit{Affiliations.txt}, which records basic information of each institution, such as their ID, standard institution name, display name, etc. The second file is \textit{PaperAuthorAffiliations.txt}, which records each author of a paper and their institution ID, original institution name (i.e., non-standard institution name), author name, and so on. The non-standard institution names are obtained by Internet crawlers or parsed from paper documents, etc. To extract our data, we first use \textit{PaperAuthorAffiliations.txt} to create mappings from the original institution names to their institution ID and then bridged on the institution IDs. We use \textit{Affiliations.txt} to find their standard institution name, which is treated as normalized institution names. 

We first deduplicate affiliation IDs by extracting all standard institution names from \textit{Affiliations.txt} and redirecting the IDs with the same standard institution name to a new ID.

As for the non-standard institution names in \textit{PaperAuthorAﬀiliations.txt}, we further preprocess these mappings to remove a bunch of noises such as unnecessary HTML tags over length names etc. However, not all of the mappings retained are correct. We found that there are a lot of conflicts within these mappings. The same original institution name could appear multiple times, and each time it is mapped to an ID, which is not necessarily always the same. Therefore, we collect all these conflicts and calculate a \textit{confidence score} for each of the ID that shows up for the original institution name. The confidence score is calculated as the portion of mapping occurrences this ID receives out from all mapping occurrences that the original institution name has. i.e. 

$$
confidence(b, i) = \frac{N_{bi}}{\sum_{j=1}^{K}N_{bj}}\in [0, 1]
$$

where $N_{bi}$ stands for the mapping occurrences that point to the $i$-th ID from the original institution name $b$. And the summation is over all the $K$ IDs that $b$ pointed to in all of its occurrences. We only retain those $b$s who have a majority votes and discard all ambiguous ones. For more details for preprocessing steos, please refer to Appendix A.

\begin{figure} [t]
	\centering 
	\subfigure[] {
		\includegraphics[width=0.465\columnwidth]{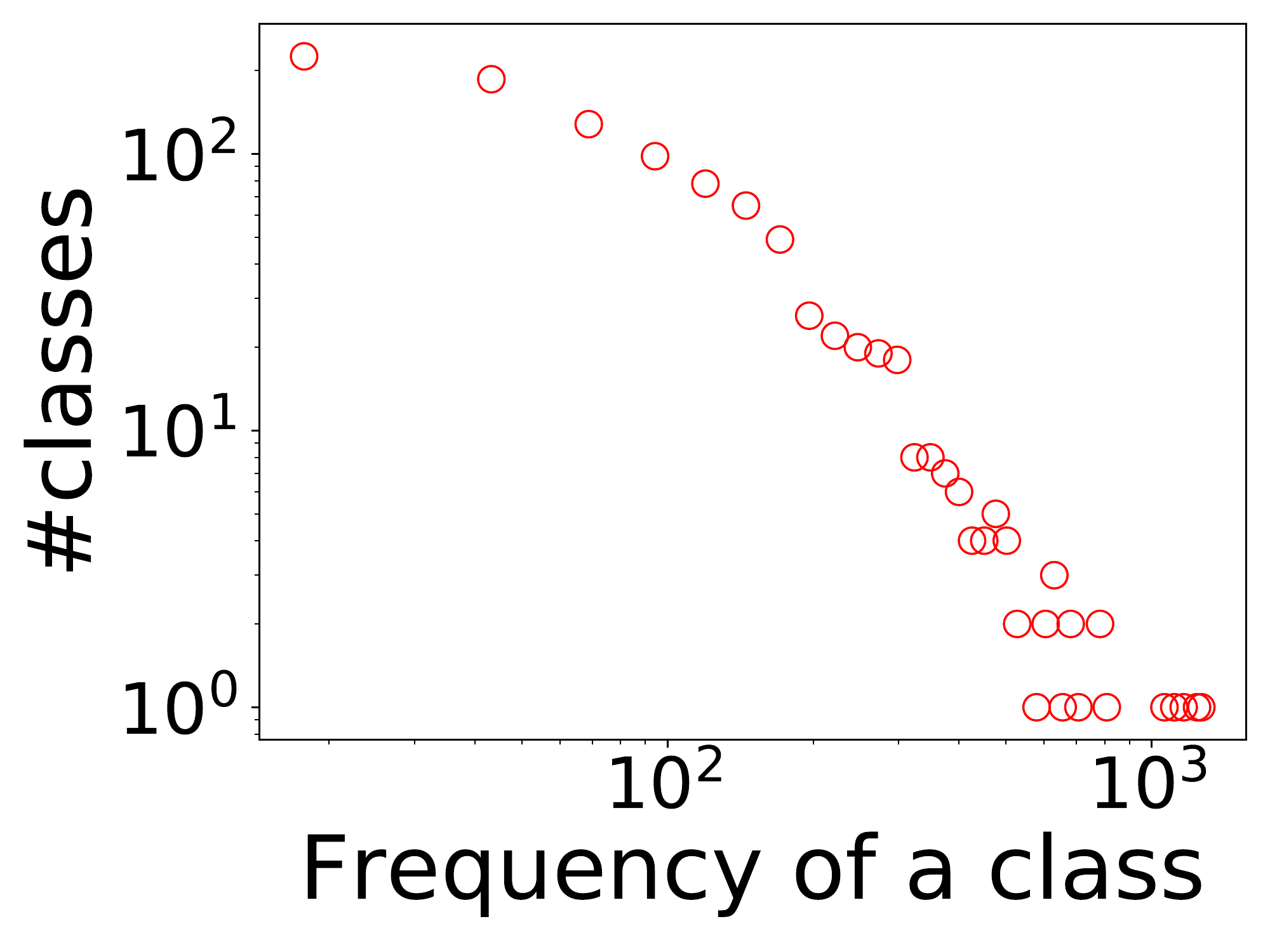}
	} 
	\subfigure[] {
		\includegraphics[width=0.465\columnwidth]{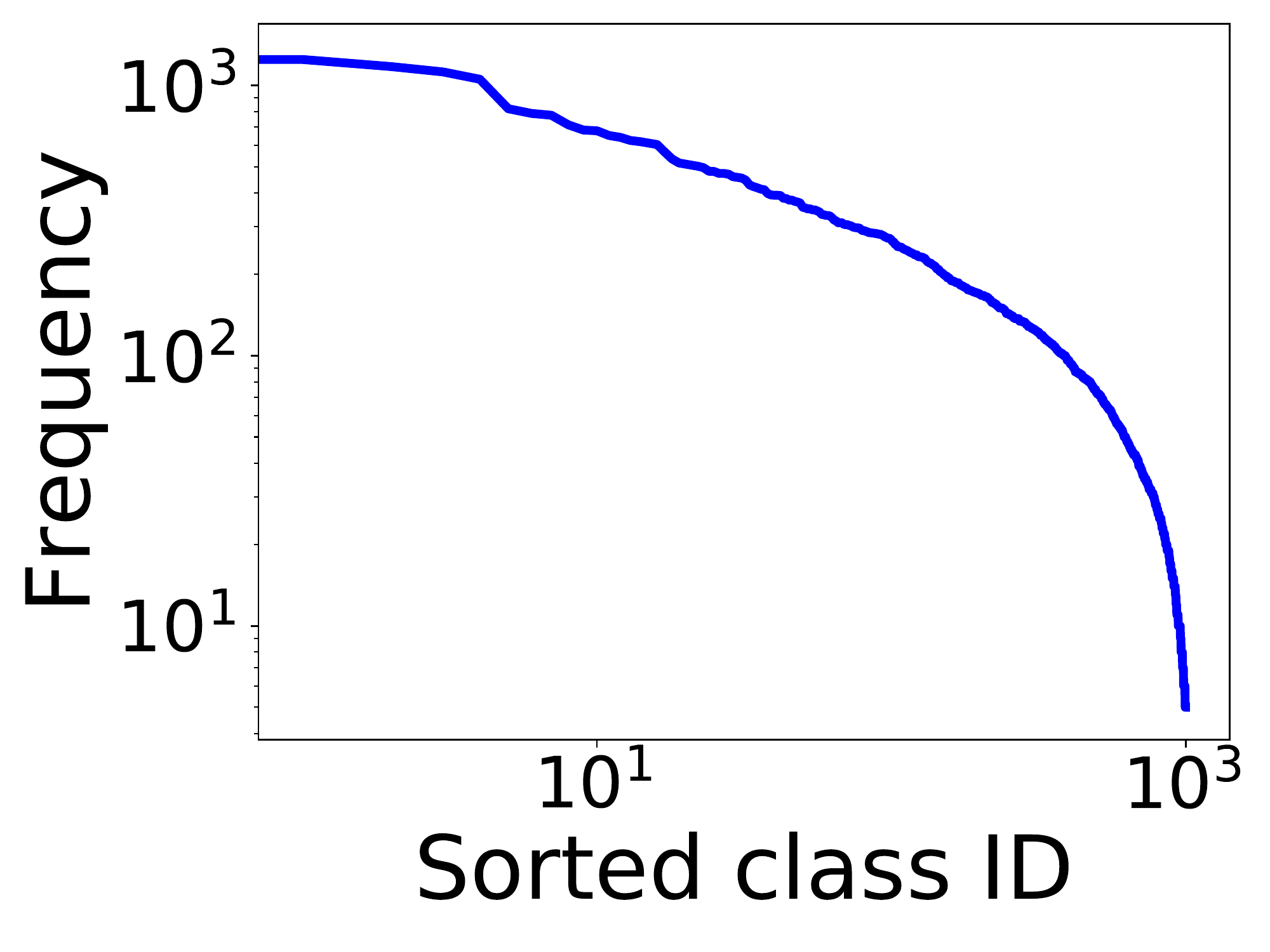}
	} 
	\subfigure[] {
		\includegraphics[width=0.465\columnwidth]{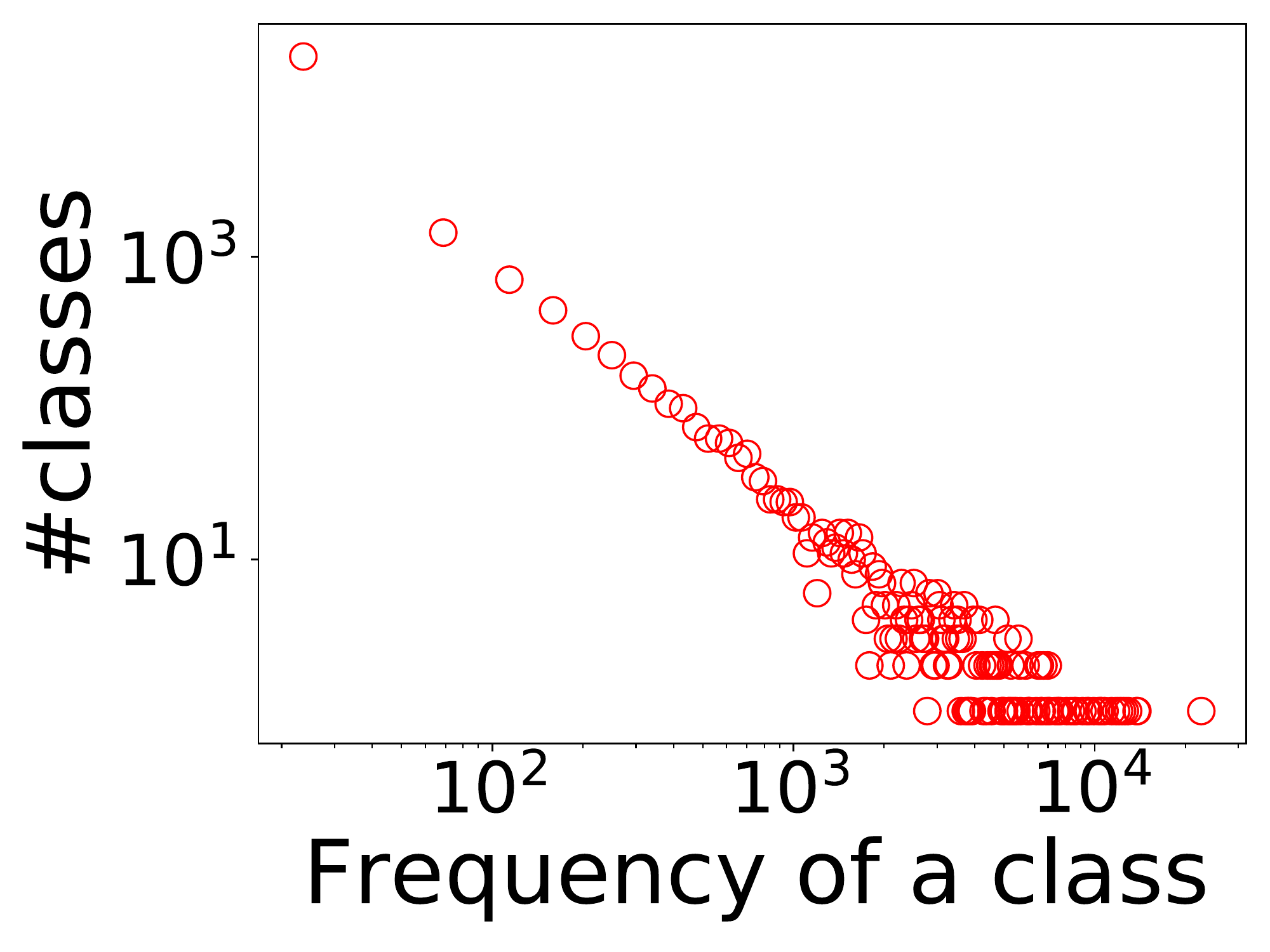}
	} 
	\subfigure[] {
		\includegraphics[width=0.465\columnwidth]{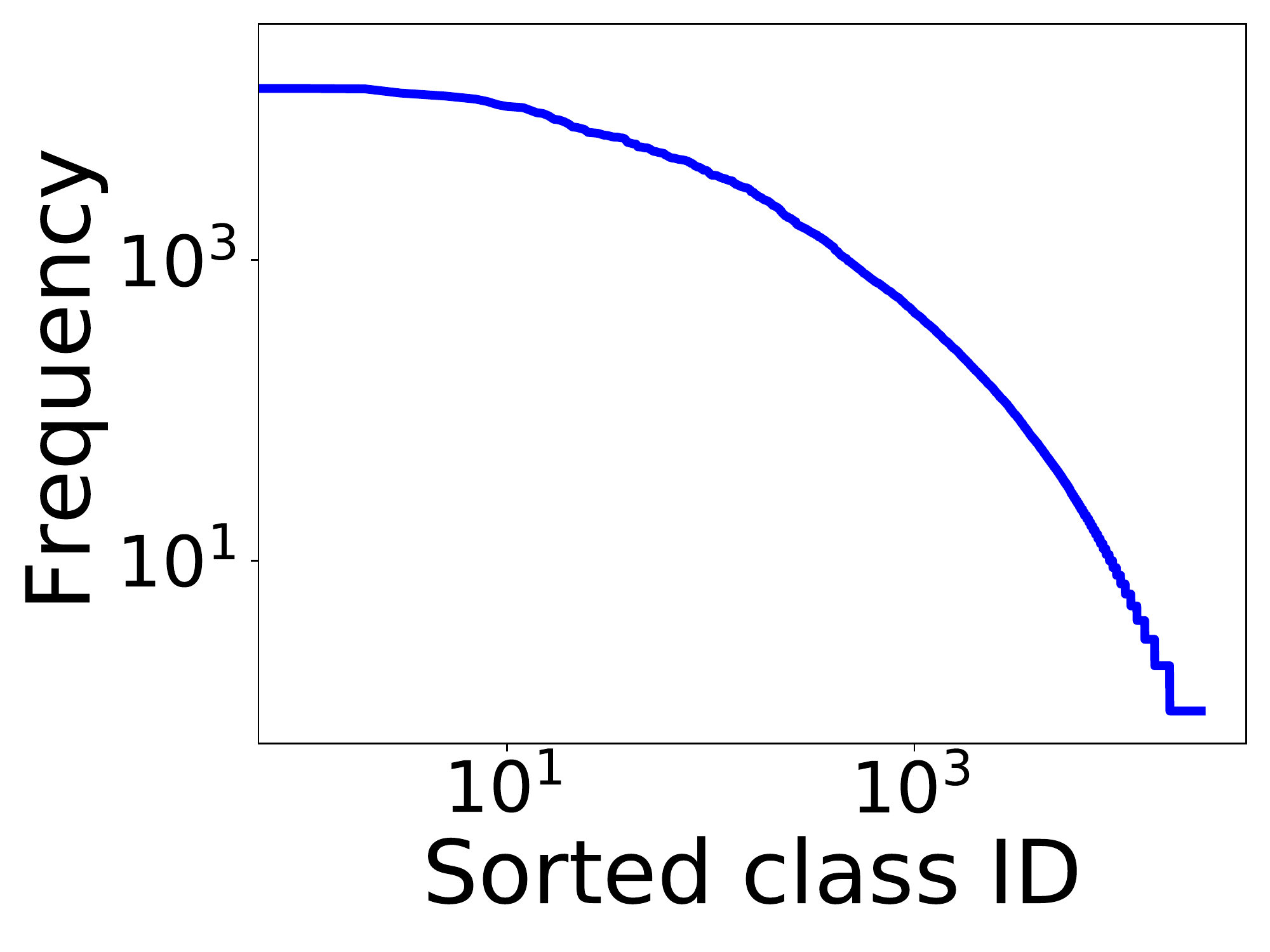}
	} 
	\caption{Comparison of the long-tailed distribution between our dataset (c and d) and ImageNet-LT (a and b). The left figures are scattered plots between the frequency of a class, i.e., counts of examples within the same class (horizontal axis) w.r.t. the number of classes with this frequency (vertical axis). To reduce scatter, we merge the data along the horizontal axis by splitting it into 500 linearly spanned bins and then summing up the class counts within each bin. For ImageNet-LT data, 50 bins are used since it has much fewer classes. The right figures are the class-frequency plot, with classes in the horizontal axis sorted by their frequency. All axes are on a logarithmic scale.}
	\label{fig:long-tail} 
\end{figure} 

\subsection{Data Filtering}
In the second data filtering stage, we focus on filtering out the redundant examples in the dataset, which effectively removes bulks of easy examples dominating the dataset. 

An illustration of these redundant examples is shown in Figure \ref{fig:connected}. An institute called ``Rowan college at Burlington County'' has 4 very close variations, all of which are the same name with minor appended abbreviations (``NJ'', ``NJ, USA'' or ``New Jersey''), or some defects related to the OCR process. (The ``1'' is from the footnote numbers in the original PDF file.) 

Redundant examples usually share a major part of their string body with others, with minor differences at the beginning or end of the string. These are easy examples that can be classified correctly with simple string matching methods. However, they constitute a big portion (over 90\%) of the dataset, which can push the classification accuracy spuriously high, submerging the more interesting, harder ones that need more sophisticated classifiers. 

Figure \ref{fig:connected} shows an example of our method of filtering out the redundancy. First, we treat every original name as a node. Second, we connect two nodes if one node is a substring of the other, thus forming an undirected graph. We then find all the connected components in the graph. For each connected component, only one node is randomly chosen to be preserved, while all other nodes within the same connected component are discarded.

\subsection{Partitioning the Dataset}
We partitioned the dataset into different subsets for training and evaluation. The \textit{open test set} was collected by randomly sampling $2\%$ of the \textit{categories}. Thus the model will not see any examples from these categories during training. For the two \textit{close test set} and \textit{valid set}, we randomly sample $2\%$ of the examples from the remaining data for each of the sets. To better handle few-shot categories, we conduct extra steps to ensure that there is at least one example in training set for each category in the test set, and the test set covers as many categories as possible. Please refer to Appendix B for detailed steps.

We further split the valid set and the two test sets into many-, medium-, and few-shot subsets by setting the threshold at 5 and 20 occurrences in the entire dataset. Since in open test sets, all examples are zero-shot examples for the training data, we name them as Frequent, Medium, and Rare subsets instead. An overall view of the whole partitioning results is shown in Table \ref{tab:dataset}.

Figure \ref{fig:long-tail} shows the distributional characteristics of our data. For comparison, same figures are plotted for the ImageNet-LT data \cite{liu2019large}, which is sampled from a Pareto distribution (Figure \ref{fig:long-tail}). We found that our dataset is distributionally different from the Pareto distribution, which is used in the manual sampling process in ImageNet-LT. Since our data is not collected through sampling, we believe this reflects a better real-world scenario.

\begin{table}[t]
\centering
\begin{tabular}{llrr}
\toprule
\multicolumn{2}{l}{}                                        & class & instance \\ 
\midrule
\multirow{4}{*}{Train set} & \multicolumn{1}{l|}{Many-shot}   & 6,693 & 2,179,840   \\ 
                          & \multicolumn{1}{l|}{Medium-shot} & 5,446  & 44,215  \\ 
                          & \multicolumn{1}{l|}{Few-shot}    & 12,990  & 14,093   \\ 
                          & \multicolumn{1}{l|}{Overall}     & 25,129 & 2,238,148  \\
\midrule
\multirow{4}{*}{Valid set} & \multicolumn{1}{l|}{Many-shot}   & 6,693 & 46,212   \\ 
                          & \multicolumn{1}{l|}{Medium-shot} & 5,446 & 5,553  \\ 
                          & \multicolumn{1}{l|}{Few-shot}    & 2,673  & 2,674   \\ 
                          & \multicolumn{1}{l|}{Overall}     & 14,812 & 54,439  \\
\midrule
\multirow{4}{*}{Close test set} & \multicolumn{1}{l|}{Many-shot}   & 6,693 & 47,153   \\ 
                          & \multicolumn{1}{l|}{Medium-shot} & 5,446 & 5,551  \\ 
                          & \multicolumn{1}{l|}{Few-shot}    & 5,445  & 5,450   \\ 
                          & \multicolumn{1}{l|}{Overall}     & 17,584 & 58,154  \\
\midrule
\multirow{4}{*}{\makecell{Open test set \\ (zero-shot)}} & \multicolumn{1}{l|}{Frequent}   & 132   & 42,985   \\ 
                          & \multicolumn{1}{l|}{Medium} & 102  & 1,043    \\ 
                          & \multicolumn{1}{l|}{Rare}    & 278   & 492     \\ 
                          & \multicolumn{1}{l|}{Overall}     & 512   & 44,520   \\ 
\bottomrule
\end{tabular}
\caption{Statistics of our dataset. Many-, middle-, and few-shot subsets are split according to a global frequency of 5 and 20. The open set subsets are also split in this way, except that the names are different since they are all zero-shot examples.}
\label{tab:dataset}
\end{table}

\section{Tasks and Evaluation Metrics}

In this section, we will introduce three tasks: close set classification, open set classification, and open set verification. Moreover, the corresponding evaluation metrics for these three tasks are also provided. 

\subsection{Closed-Set Classification (CSC)} \label{csc}
This task is mostly the same as the canonical classification task. As the test set only contains classes available in the training set, the model assumes that all test samples are from one of the known classes. We use 2 standard evaluation metrics, i.e., the accuracy (a.k.a. micro F$_1$) and macro F$_1$ measure. 

\subsection{Open-Set Classification (OSC)} \label{osc}
In this task, the model is asked to tell if a given sample belongs to an unseen class or not. To comply with all the previous methods in the literature, we conduct this task on top of the model trained in the CSC task setting. We set a threshold for the largest probability over all known classes and classify the example as an unseen class if the largest probability over all known classes is smaller than the threshold.
Since there is a sensitivity-specificity trade-off in determining the threshold, we set the evaluation metric as comparing the ROC (receiver operating characteristic) curve between different methods. 

\subsection{Open-Set Verification (OSV)} \label{osv}
In addition to the open set classification task, we propose a new open set verification task, which provides a more fine-grained evaluation of the model's performance on the unseen data. In this task, we first sample a pair of examples in the open set, and then the model is asked to tell if the two provided samples belong to the same unseen class or not. It reflects whether the model could transfer its learned knowledge in telling apart examples from unseen classes.  
To achieve this, we manually sample a fixed test set of name pairs from the open test set, controlling the number of positive (belonging to the same class) and negative pairs (belonging to different classes) to be balanced. We can compare different models on the same testbed by applying the same fixed test set to all models. The prediction accuracy is used as the evaluation metric.

\section{Methods}
In this section, we reproduce five baseline methods from the literature and then describe our proposed method. 

\subsection{Baselines}
The baseline methods used in our work include two basic machine learning model which are Naive Bayes \cite{maron1961automatic} and Fasttext \cite{joulin2016bag} classifier, two search-based methods which are sCool \cite{jacob2014scool} and CompanyDepot V1 (CDv1) \cite{liu2016companydepot}, and one deep learning model which is BERT \cite{DBLP:conf/naacl/DevlinCLT19}. All of these methods are used in the CSC task, while only Naive Bayes and BERT are used in OSC and OSV tasks due to the nature of these methods. C.f. Appendix C for more descriptions of these models. 

\subsection{Our Method}

\begin{figure}[t]
    \centering
    \includegraphics[width=0.48\textwidth]{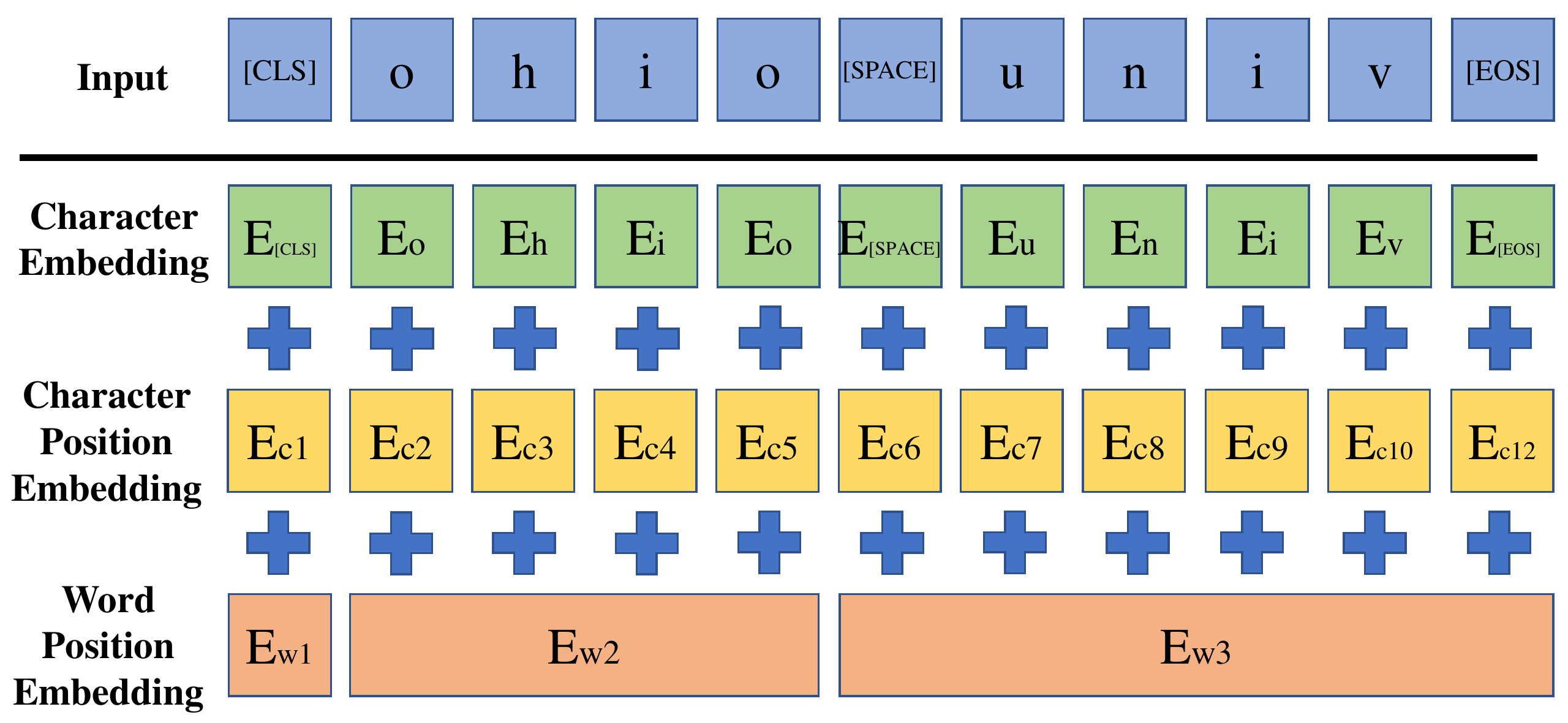}
    \caption{Overview of the input embedding layer. All three embedding components are learnable parameters, and element-wise added together. Word position embedding are aligned according to the word boundaries. Apart from the character vocabulary, 4 extra special marks: \texttt{[CLS], \texttt{[EOS]}, \texttt{[PAD]}, \texttt{[MASK]}}, which are used in input embedding layer. When constructing character embedding, \texttt{[CLS]} and \texttt{[EOS]} are used at the beginning and end of the sequence, respectively. \texttt{[PAD]} is used for padding in mini-batch training, and \texttt{[MASK]} for pretraining. }
    \label{fig:input}
\end{figure}

\begin{figure}[t]
    \centering
    \includegraphics[width=0.99\linewidth]{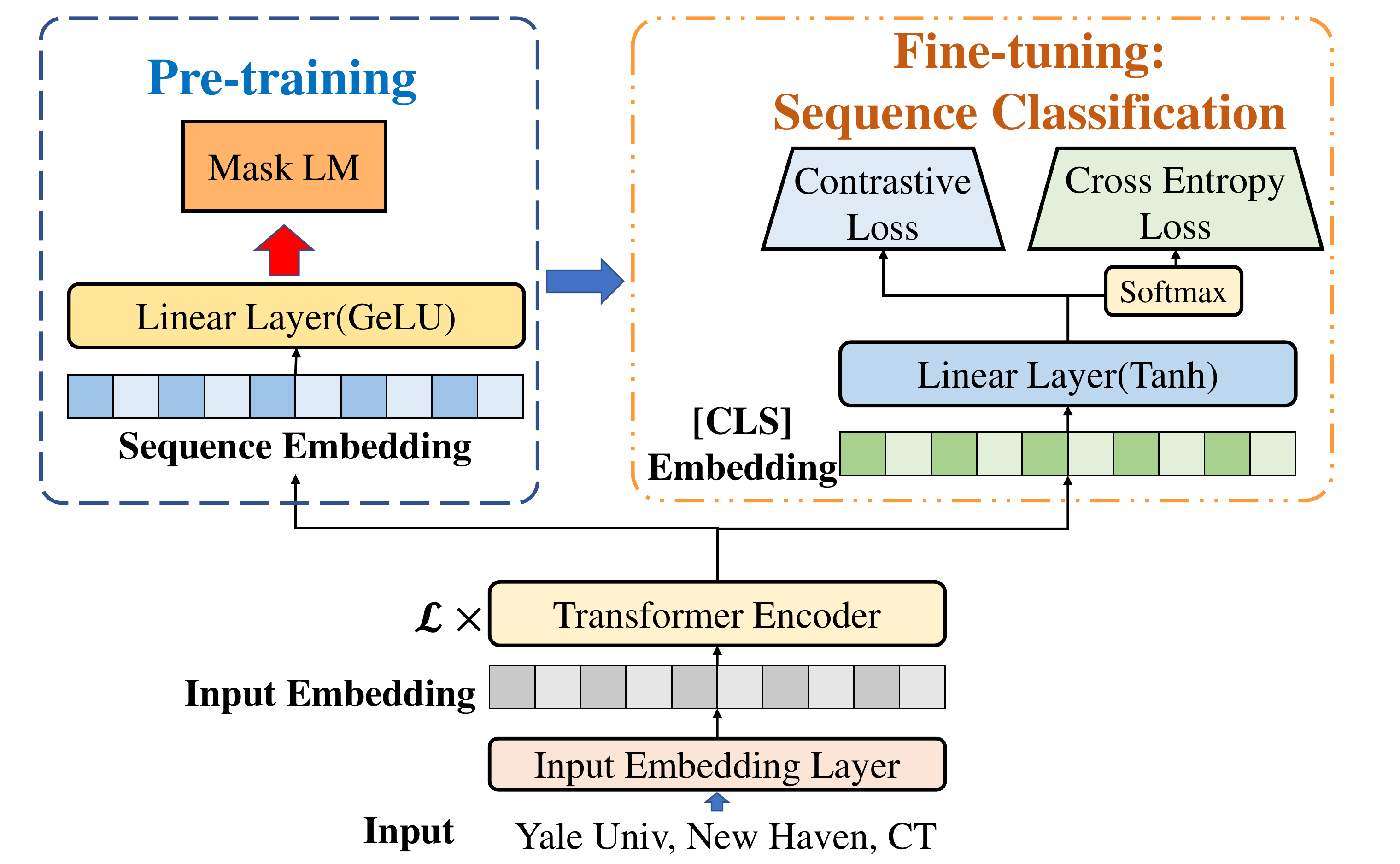}
    \caption{Overview of our model architecture.}
    \label{fig:model}
\end{figure}

In this paper, we introduce a special character-level BERT-based model, which is first pretrained from scratch on the institution name corpus and then fine-tuned as a sequence-level classifier. The overview of the model architecture is shown in Figure \ref{fig:model}.

Since institution names are usually shorter than natural sentences and contain many out-of-vocabulary (OOV) words, we modify the original BERT model to character level. The input embedding is modified to consist of 3 parts, i.e., character embedding, character position embedding, and word position embedding. We also remove the segment embedding of the original BERT because we do not input a pair of sentences. To still give indications of word boundaries, we add word position embedding to the input embedding. The overview of the composition of input embedding is shown in Figure \ref{fig:input}. 

In the pretraining stage, we only use the masked-language modeling (MLM) task, except that masks are all at the character level. In order to use as much data as possible during pretraining, the model is trained in all institution name records before the data cleaning step. In the fine-tuning stage, we choose sequence-level classification as the downstream task. A softmax layer is added after the output of \texttt{[CLS]} tokens to predict the corresponding class. 

\begin{table*}[htbp]
\centering
\begin{tabular}{cc||ccccc|cc|c}

\toprule
  \multicolumn{2}{c||}{Metrics \textbackslash Methods}  & Naive Bayes & sCool & CD-V1  &FastText  &  BERT      & BERT+RS            & BERT+RS+FP      & Ours'          \\ \midrule
\multirow{2}{*}{Overall}            & Accuracy          & 72.20       & 76.72 & 79.97 &74.93   & 83.30         &83.40            &\textbf{84.00}    & 83.73      \\ 
                                    & Macro - $F_1$     & 50.20       & 52.41 & 59.64  &44.38  & 62.79         &63.98            & 65.14        & \textbf{65.90}      \\ \midrule 
\multirow{2}{*}{Many-}          & Accuracy          & 78.74       & 84.41 & 86.32 &85.08   &\textbf{90.01} &89.67            &\textbf{90.01}& 89.12               \\ 
                                    & Macro - $F_1$     & 69.80       & 74.09 & 76.88 &69.84    &82.09          &82.28            &\textbf{82.43}& 81.66               \\ \midrule
\multirow{2}{*}{Medium-}        & Accuracy          & 56.58       & 56.53 & 62.98  &49.75  & 67.80         &69.17            & 69.93        & \textbf{71.68}      \\ 
                                    & Macro - $F_1$     & 55.82       & 55.97 & 62.56 &38.56   & 67.55         &68.82            & 69.94        & \textbf{71.05}      \\ \midrule
\multirow{2}{*}{Few-}           & Accuracy          & 31.52       & 30.78 & 42.29  &12.99  & 40.99         &43.66            & 46.26        & \textbf{49.36}      \\ 
                                    & Macro - $F_1$     & 31.16       & 30.52 & 42.03  &8.69  & 40.75         &43.33            & 45.99        & \textbf{49.06}      \\ 
\bottomrule

\end{tabular}
\caption{The performance of different methods in CSC task. "RS" is short for "resampling". "FP" is short for "further pretrain"."Many-" is short for "Many shot" as well as "Medium-" and "Few-".}
\label{tab:result-s}
\end{table*}

We also use a resampling strategy in our method, which adjusts the sampling probability for each class according to its frequency. The probability $p_j$ of sampling a data point from class $j$ is given by:
\begin{equation}
p_{j}=\frac{n_{j}^{q}}{\sum_{i=1}^{C} n_{i}^{q}}
\end{equation}
where $n_j$ denotes the number of training examples for class $j$. We choose $q < 1$ in order to avoid medium and few shot categories being submerged by many-shot ones.

Specifically for the OSV task, we introduce an additional loss term during fine-tuning based on contrastive learning \cite{hadsell2006dimensionality}, which pushes two feature vectors close if they belong to the same class and pushes them away from each other otherwise. We follow the convention of using the output vector of \texttt{[CLS]} as the feature vector of the whole sequence, i.e., the institution name. Formally, the loss is defined as

\begin{equation}
  \label{eq:contrastive}
  l(i, j) = y_{ij} d_{ij}^2 + (1-y_{ij}) [\alpha - d_{ij}]_+^2~,
\end{equation}

where $d_{ij}$ means the Euclidean distance between the feature vector of a sample $i$ and $j$, and $y_{ij}$ means whether sample $i$ and $j$ belong to the same class or not. $\alpha$ is a tunable hyper-parameter. Fine-tuning by the contrastive loss will gain a feature space with meaningful distance metric, where the distance indicates their relativity. 
Since the scale of the training set is very large, we sample a part of the training data as anchors. For the OSV task, the distance can be used to determine if two given names belong to the same unseen class or not. 

\section{Experiments}
The experiment section consists of four parts, including evaluations of the four baseline models and our proposed model for the three tasks that come with our dataset, as well as some ablation studies on our proposed model for the dataset. 
\subsection{Closed-Set Classification}

\begin{figure*} [ht]
	\centering 
	\subfigure[Overall] {
		\includegraphics[width=0.475\columnwidth]{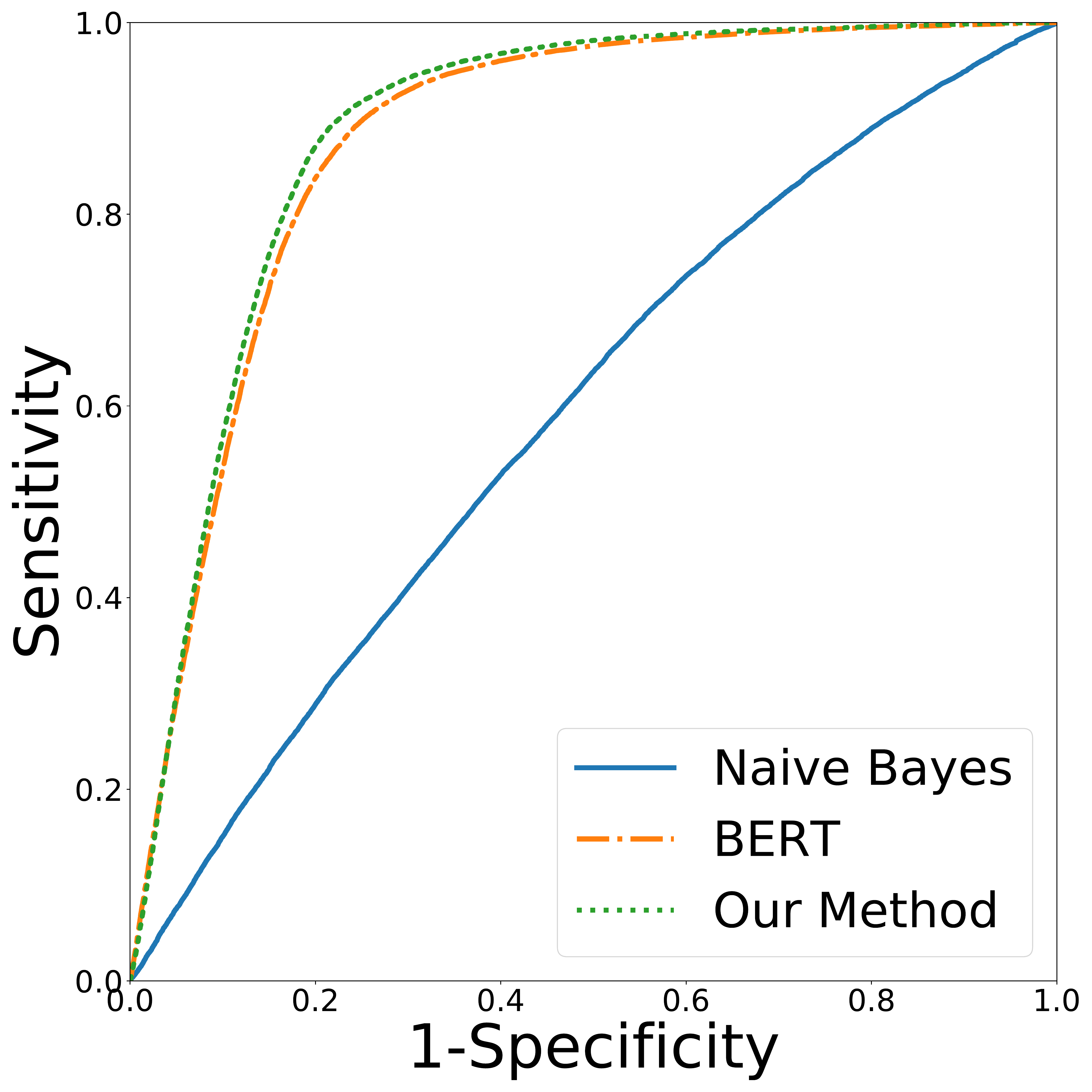}
	} 
	\subfigure[Frequent subset] {
		\includegraphics[width=0.475\columnwidth]{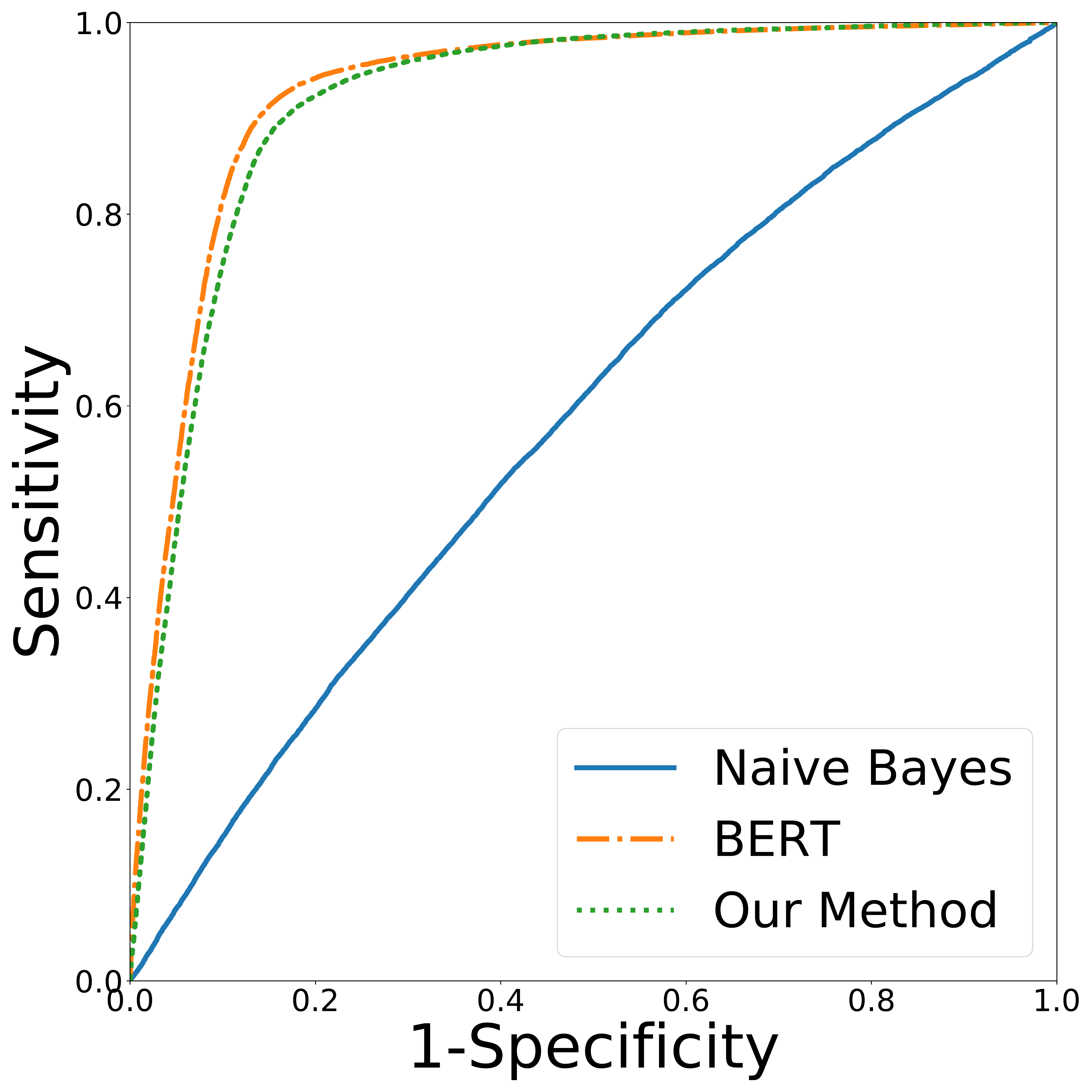}
	} 
	\subfigure[Medium subset] {
		\includegraphics[width=0.475\columnwidth]{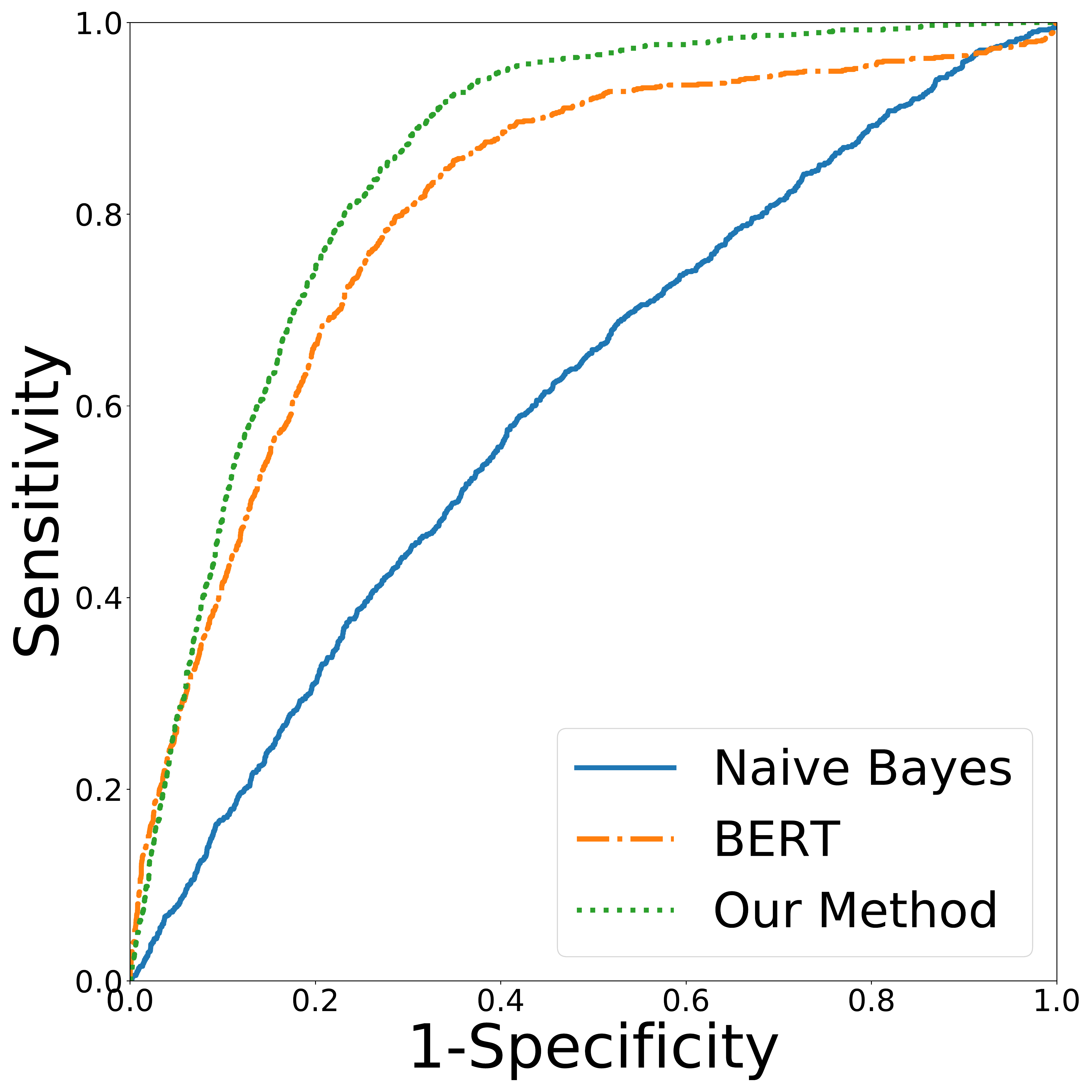}
	} 
	\subfigure[Rare subset] {
		\includegraphics[width=0.475\columnwidth]{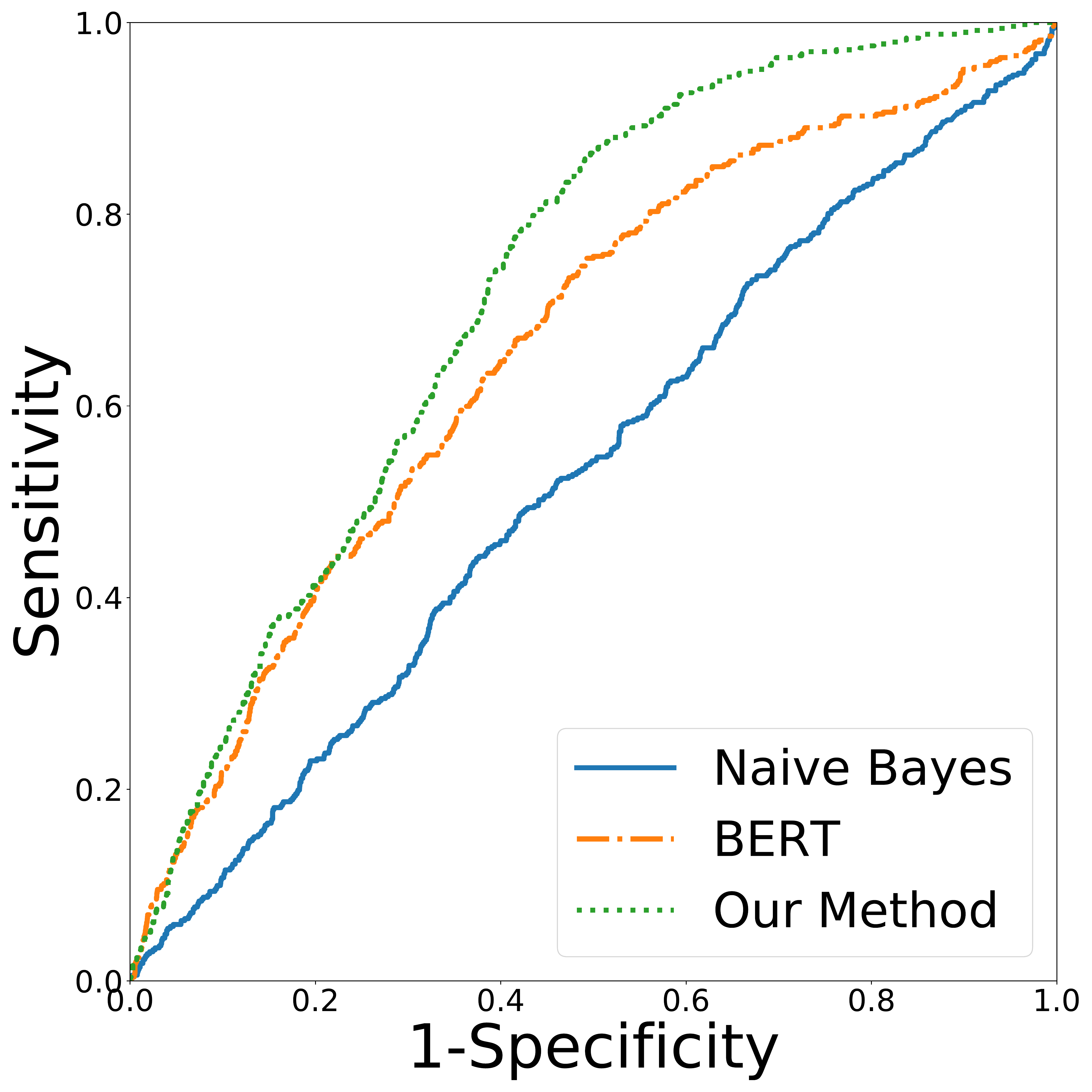}
	} 
	\centering
	\caption{ The performance in open set for three methods. (a) shows the overall performance while (b)(c)(d) shows the performance in frequent, medium and rare subset.} 
	\label{fig:osc_result} 
\end{figure*}

We report performances of all aforementioned methods on our close test set (Table \ref{tab:result-s}). 
Apart from the observation that almost all baseline methods perform far worse on the few-shot test set than on the many-shot test set, our BERT-based method significantly outperforms all other baseline methods of various types, which indicates the effectiveness of pretraining on this task.

It is also worth noting that our character-level pretrained BERT model further outperforms the original BERT fine-tuned on the medium- and few-shot subsets. However, our method's performance is inferior to BERT on many-shot subsets, indicating conflicts and trade-offs between the performances of many-shot and few-shot subsets, thus leaving space for future research.

\subsection{Open-Set Classification}
We report all methods' performance except for two retrieval methods that are not suited for this task. Using a mixed dataset which consists of the whole open set and an equal number of samples drawn from the close set, we plot ROC curves for these methods on each of the subsets (Figure \ref{fig:osc_result}).

From Figure \ref{fig:osc_result}, we can tell that our proposed model has a very similar performance to the original BERT method in the overall test set and many-shot subset. However, our proposed model achieves significantly better performance in medium-shot and few-shot subsets than other approaches. 

\subsection{Open-Set Verification}
Except for the two retrieval methods that are not suited for the verification task, we report the other two baselines and our method for this task. For the model fine-tuned with contrastive loss, we simply use the L2 distance between the feature vectors of samples in a pair as the distance measure. For other models, we first calculate the output distribution over all known classes for each input in the same pair and then calculate the Jensen-Shannon divergence (JSD) between them as their distance measure. 
Table \ref{tab:task3} shows the accuracy of different methods. One can observe that training with the contrastive loss significantly increases the model's ability to tell apart different samples, even if they are from unseen classes. 

\begin{table}[t]
    \centering
    \begin{tabular}{cc}
    \toprule
       Methods  &  Accuracy \\ \midrule
       Naive Bayes  &  69.12   \\
       BERT  &  78.88 \\
       Our Method  &  \textbf{79.79}  \\
      \bottomrule
    \end{tabular}
    \caption{Test accuracy on OSV task.}
    \label{tab:task3}
\end{table}

\subsection{Ablation Study}
In this section, we describe two more experiments we did as an ablation study to reflect the importance of several components in our pretrained character-level model. 

\textbf{Character-level v.s. Word-level Pretraining}
Although our model significantly surpasses the baseline BERT model in the aforementioned experiments, one may argue that it may be the resampling strategy in our model that made this difference. We then train another BERT model with resampling during fine-tuning (marked as BERT+RS in Table \ref{tab:result-s}). All other settings are set to be the same as our model, making the only difference being that our model is a character-level model, while BERT is at subword level. It can be seen that with resampling, the performance on the few-shot subset improved, suggesting the resampling strategy's effectiveness. However, the BERT+RS model is still inferior to our model, showing the necessity of using a character-level model. 


\textbf{Pretraining from Scratch v.s. Further Pretraining}
Considering that BERT is pretrained on a general corpus, while our model is pre-trained solely on the proposed dataset, we add another stage of training in the BERT baseline, which is further pretraining the original BERT model on our dataset before fine-tuning (marked as BERT+RS+FP in Table \ref{tab:result-s}). We can figure that the BERT+RS+FP improves over BERT+RS by a large margin, suggesting that our dataset does have a big difference from the general corpus used by BERT pretraining. Compared to our model, which is pretrained from scratch at the character level on our dataset, the model outperforms in the many-shot subset while still lags a significant gap in the medium and few shot subsets. 

The above two experiments show the necessity of using a character-level model, as well as suggest the difference between our dataset and a general natural language corpus. We conjecture that this is due to the enormous place names or surnames in various languages across the world that results in a vocabulary that the subword vocabulary of a general corpus could not meaningfully cover. The character-level model could better handle other challenges in our dataset, such as the character-level typo errors caused during PDF parsing or OCR procedures.

\section{Conclusion}
\label{ssec:layout}

In this paper, we present a large-scale institution name normalization dataset, which exhibits a long-tailed distribution in over 25k classes. We reproduce different types of publicly available methods for this task, providing a fair comparison between different methods. Also, a new BERT-based model, as well as a contrastive loss for this task, is proposed, which outperforms previous methods, setting a strong baseline for the dataset.

Compared to other datasets focused on the long-tailed phenomenon, our dataset has one order of magnitude more training data than the largest existing long-tailed datasets and is naturally long-tailed rather than manually resampled. We believe it provides an important and different scenario to study this problem. We hope our study can pave the way toward long-tailed text classification.

\section*{Acknowledgement}
This work was sponsored by the National Natural Science Foundation of China (NSFC) grant (No.
62106143), and Shanghai Pujiang Program (No. 21PJ1405700).

\clearpage

\vfill\pagebreak



\bibliographystyle{IEEEbib}
\bibliography{Template}

\clearpage
\appendix

\section{Detailed Data Cleaning Procedures} \label{appendixA}



Our preprocessing steps are briefly summarized as follows:

\begin{enumerate}
    \item Replace meaningful non-ASCII characters with the corresponding ASCII characters.
    \item Lowercase all letters.
    \item Converts non-display ASCII characters in the organization name to whitespace characters.
    \item Converts HTML tags to whitespace characters.
    \item Converts characters that are not in Table \ref{tab:chars} to whitespace characters.
    \item Replace consecutive whitespace characters with a single whitespace character.
    \item Remove whitespace characters at the beginning and end of the institution name.
    \item Discard this name if the number of letters is less than 2 or more than 254.
    \item Discard this name if the number of words is more than 63.
\end{enumerate}


\begin{table}[h]
    \centering
    \begin{tabular}{l c l }
    \hline
    category&number& comment\\
    \hline
      lowercase  &  26  &  \\
        digital &  10  &  \\
      whitespace  &  1  &  \\
        punctuation &  17  &  ”\#\&’()*,-.:;@ $\backslash$ \_\{\}\\
    \hline
    \end{tabular}
\caption{Characters ratained in preprocess.}
\label{tab:chars}
\end{table}

\section{Additional Steps in Partitioning the Dataset} \label{appendix_sampling}
We partition the dataset following these steps:
\begin{enumerate}
    \item Randomly samples 2\% of the categories and take out all samples from these categories as the open test set.
    \item Randomly samples 2\% of the mappings as the closed test set. And the remaining data is the training set.
    \item If all varients of an institution were divided into train set, pick one variant and move it to the test set. This step is to make the test set can cover as many categories as possible.
    \item After the previous step is done, if all variants of an institution were divided into test set, pick one variant and move it to the train set. This step is to ensure that every class has been seen by the model in the train set.
\end{enumerate}
The step used to partition the train set and the validation set is the same as the above method. It just partitioned the train set, which was already partitioned in the last step.

\section{Details for Experiment Implementation}
\label{bertdetail}

\subsection{Details for Experiment Setting for Baseline}
\subsubsection{Naive Bayes}
\label{nb}
Our first baseline is a basic machine learning model, the Naive Bayes classifier \cite{maron1961automatic}. We tokenize the institution names and clear stop words to form bag-of-words features, which are inputs to the Naive Bayes classifier.

It is straightforward to apply it to the CSC task. And for OSC, we follow the steps in the open set classification Section to get its ROC curve. 

For OSV task, we feed the two names in a pair, i.e., $b_1$ and $b_2$, separately as input to the model, and get their output distribution $p_1$ and $p_2$.
Then we calculate the Jensen-Shannon divergence between them $JSD(p_1||p_2)$ as their dissimilarity. 
We first evaluate it on the validation set to determine the best threshold and then report the performance on the test set.

\subsubsection{FastText}
Our second baseline is a deep learning method. We use FastText \cite{joulin2016bag}
, which is a simple but strong baseline in text classification. It consists of BoW features and a linear classifier.

For the three tasks, we follow the same steps as in Naive Bayes.

\subsubsection{sCool}
Our third baseline is sCool \cite{jacob2014scool}, which is one of the earlier studies on the task of normalization of institution names.
Different from pure classification models, it first initializes a database with the given data and then retrieves over it. Also, sCool obtains better results by refining the search results in the retrieval step. In our implementation, we use ElasticSearch\footnote{https://www.elastic.co/elasticsearch/} indexed with training data. Due to its algorithmic nature, in our experiments, we only apply sCool on the CSC task.

\subsubsection{CompanyDepot} \cite{liu2016companydepot} proposed CompanyDepot v1 (CDv1) system
\footnote{Note that there is a more recent version for this method, named CDv2, which defines the tasks of cluster level and entity level, adds multiple data sources to construct two Lucene search engines, uses query expansion to search in retrieval step, and adds several extra features in reranking step. However, due to a lack of detailed information and extra data, it is not possible to exactly reproduce their results. So we are using CDv1 as our baseline. } 
for company name normalization, which we take as our third baseline. The whole system consists of four steps: index, retrieval, rerank and validate. In the index step, a search engine is built for retrieval. In the retrieval step, it uses a variety of rules to extract candidates from the search results. In the rerank step, a set of 45 manually designed features are used for learning to rank algorithm. In the validate step, it uses a LibSVM as a binary classifier to determine whether it is an unseen class or it should output the top-ranked class as a result. CDv1 uses a machine learning approach compared to sCool, which significantly improves accuracy.

In our implementation, we also built an ElasticSearch engine indexed with training data. To train the learning-to-rank algorithm and LibSVM classifier, we split out about 20K data from the training set. Again, because of its algorithmic nature, we only use it on CSC task.

\subsubsection{BERT}
BERT \cite{DBLP:conf/naacl/DevlinCLT19} is a natural language processing pretraining technique based on Transformer architecture. Pretraining tasks for BERT include masked language modeling (MLM) and next sentence prediction (NSP). The former is used to predict the masked token given its context, while the latter is used to determine whether two sentences are continuous.
In our experiment, we fine-tune over the pretrained $\rm{BERT_{BASE}}$ model on our dataset, with the sequence-level classification setting as specified in the paper of BERT as our baseline.

\subsection{Details for Experiment Setting for Our Method}

We trained our models on one Linux 18.04 machine with 4 GeForce RTX 3090. We use Pytorch with version 1.7 to implement all the algorithms. The pretraining task was trained for about 220 hours. Fine-tuning for classifier and Fine-tuning with contrastive loss were both trained for about 20 hours.

The character vocabulary size is 58, including the 54 characters shown in Table \ref{tab:chars} and 4 extra special marks: \texttt{[EOS]}, \texttt{[MASK]}, \texttt{[PAD]}, \texttt{[CLS]}, which are used in input embedding layer. When constructing character embedding, we put a \texttt{[CLS]} mark at the beginning of the sequence, which will collect the information of the whole sequence. We put a \texttt{[EOS]} mask at the end of the sequence. The \texttt{[PAD]} mark is used for padding the sequences with different lengths to the same length when we train the model using a mini-batch. The \texttt{[MASK]} mark is used for the masked-LM task in the pretraining stage. The max length of character position embedding is 256, and the max length of word position embedding is 64.

We use the same BERT architecture of $\rm{BERT_{BASE}}$ (L=12, H=768, A=12). We use a model with 12 layers transformer encoder, 768 hidden layer dimension, and 12 heads self-attention (i.e. L=12, H=768, A=12).


In the pretraining stage, the batch size is 32, and the learning rate is lr=5e-5. Moreover, we create a schedule with a learning rate that decreases linearly from the initial learning rate set in the optimizer to 0 after a warmup period, during which it increases linearly from 0 to the initial learning rate set in the optimizer. The learning rate changes linearly every 30 mini-batches (called a step). The whole pretraining stage contains 160k steps. We use about 70k steps checkpoint to do the downstream task for it has been converged. The mask strategy for masked-LM is the same as the origin BERT. It is worth noting that masking may replace some space with a non-space character or otherwise, so the word position embedding may change after masking.

In the fine-tuning stage of the BERT classifier, the batch size is 32, and the learning rate is lr=5e-5. We create a schedule with a constant learning rate preceded by a warmup period during which the learning rate increases linearly between 0 and the initial learning rate set in the optimizer. The learning rate will increase from 0 to max value and keep it to end. The warmup stage has 1000 steps. The total training stage contains 50000 steps. Because of the imbalanced training data, we use a special data picking strategy in training. For each data picking operation, we first pick a class. The probability of picking a class is determined by the resampling parameter. Secondly, we pick an institution name belonging to this class uniformly. We repeat this picking operation $bs$ times ($bs$ is the batch size) and pack them as a mini-batch.

In the fine-tuning stage of BERT with contrastive loss, the batch size is 16, and the learning rate is 2e-5. The learning rate scheduler and the number of training steps are the same as the BERT classifier. 
For loss calculation, we use a hard sample mining strategy. We calculate the L2 distance of all pairs in a mini-batch, then pick the top 4 positive pairs (i.e., two samples are in the same class) with maximum L2 distance and the top 8 negative pairs with minimum L2 distance to calculate the contrastive loss. We use the kNN classifier for the CSC task. Because of the bulky size of the training data, we sample a part of the training data as anchors for the kNN classifier to avoid taking too much time and memory when classifying.

\end{document}